\documentclass{article}

\PassOptionsToPackage{sort&compress}{natbib}
\usepackage[preprint]{corl_2026} %

\usepackage{booktabs}
\usepackage{array}
\usepackage{multirow}
\usepackage{graphicx}
\usepackage{amsmath}
\usepackage{amssymb}
\usepackage{subcaption}
\usepackage{hyperref}
\usepackage[table]{xcolor}
\usepackage{wrapfig}
\usepackage{afterpage}
\usepackage{placeins}

\newcommand{\method}{\textsc{CAP}}

\title{CAP: Continuously Adaptive Perception-Blind Humanoid Locomotion via Learned Denoising}

\author{%
  Hongjin Chen$^{1,2}$\quad Zijun Xu$^{1,3}$\quad Shihao Ma$^{1}$\quad Yi Zhao$^{1}$\\
  \bfseries Xilai Liu$^{4}$\quad Ke Ma$^{1,2}$\quad Wei Zhang$^{4}$\quad Chunyang Xie$^{4}$\\
  \bfseries Pengfei Li$^{2}$\quad Jieru Zhao$^{5}$\quad Wenchao Ding$^{1,2,*}$\\[3pt]
  {\normalfont $^{1}$Fudan University\quad $^{2}$TARS Robotics\quad $^{3}$Shanghai Innovation Institute}\\
  {\normalfont $^{4}$Harbin Institute of Technology\quad $^{5}$Shanghai Jiao Tong University\quad $^{*}$Corresponding author}\\[4pt]
  {\normalfont\href{https://hoshi-no-ai.github.io/CAP/}{\texttt{hoshi-no-ai.github.io/CAP/}}}%
}

\hypersetup{%
  pdfsubject={Accepted at the Conference on Robot Learning (CoRL), 2026},
  pdftitle={CAP: Continuously Adaptive Perception-Blind Humanoid Locomotion via Learned Denoising},
  pdfauthor={Hongjin Chen, Zijun Xu, Shihao Ma, Yi Zhao, Xilai Liu, Ke Ma, Wei Zhang, Chunyang Xie, Pengfei Li, Jieru Zhao, Wenchao Ding}%
}

\begin{document}
\maketitle

\begin{nolinenumbers}
\begin{center}
  \captionsetup{type=figure}
  \vspace{-10pt}
  \includegraphics[width=\linewidth]{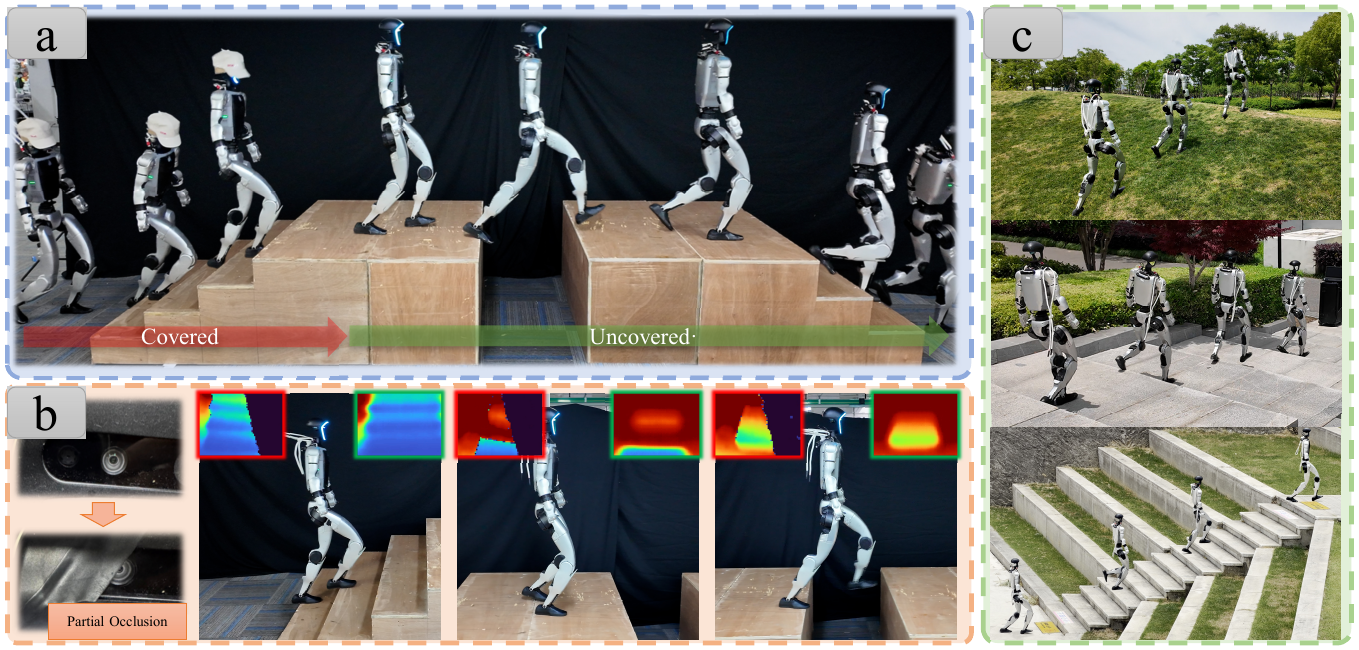}
  \captionof{figure}{\method{} enables a single humanoid policy to traverse complex
    terrain while adapting continuously under changing and corrupted
    perception. (a) Mixed-terrain cover--uncover episode. (b)
    Partial-occlusion examples across terrains, with raw depth inputs (left, red)
    and world-model reconstructions (right, green) shown side by side. (c) Outdoor deployment streams on grass and
    stairs.}
  \label{fig:teaser}
\end{center}
\end{nolinenumbers}

\begin{abstract}
    Humanoid locomotion across complex terrain demands forward-looking
    exteroception to anticipate obstacles, yet this signal is unreliable
    in real-world deployment, failing partially and intermittently. Existing
    perceptive policies often assume that depth observations remain clean
    and in-distribution, while recent attempts to unify perceptive and
    blind control typically route or switch between separate sub-policies,
    leaving recoverable information in partially corrupted depth
    unexploited. We instead
    propose \method{}, a single-stage humanoid locomotion policy that
    recovers this signal with a perceptive world-model encoder trained
    as a learned denoiser to reconstruct clean depth from a corrupted
    input, together with a co-active proprioceptive variational encoder that
    supplies depth-free body-state information. A coupled
    training recipe pairs a depth-noise curriculum on the world-model input
    with world-model feature dropout on the policy-facing latent, exposing
    the policy to failures across the entire perception-quality spectrum.
    In simulation, \method{} matches or improves upon perceptive baselines
    when depth remains informative, and degrades more smoothly than a
    binary-switching baseline as perception worsens. On the Unitree G1, controlled trials and
    indoor--outdoor deployments demonstrate perception-robust locomotion
    under intermittent occlusion, real-sensor corruption, and outdoor depth
    artifacts.
\end{abstract}

\keywords{humanoid locomotion, world models, perception robustness}

\section{Introduction}
\label{sec:intro}

Humanoid robots are increasingly expected to walk over unstructured
terrain, including stairs, gaps, and
obstacles~\citep{radosavovic2024realworld, wang2026apex, wu2026php,
zhang2026wholebody, dai2026planc}, where reacting to the ground
underfoot is not enough and the policy must look ahead.
Forward-looking exteroception, such as onboard depth, supplies
this anticipation, but the signal is unreliable in deployment: motion
blur, self-occlusion, and intermittent dropouts can corrupt the depth
stream. Rather than being all-or-nothing, these corruptions are often
graded and transient: depth may be partially degraded in one moment and
usable again shortly after. A locomotion policy for the real world must
therefore stay competent across the entire perception-quality spectrum,
from clean depth through partial corruption to complete loss.

\begin{figure}[t]
  \centering
  \includegraphics[width=\linewidth]{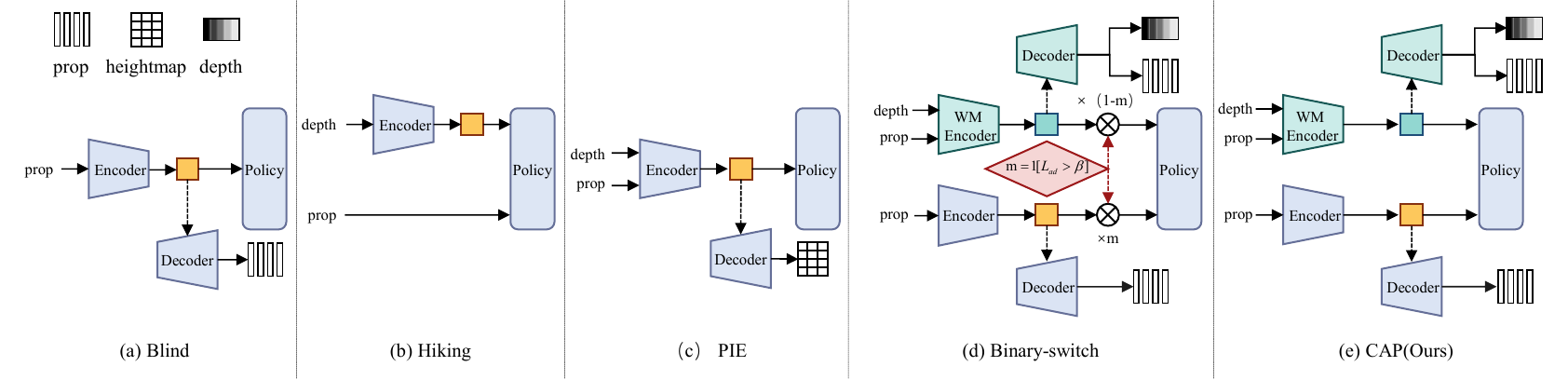}
  \caption{\textbf{Paradigms for perception-robust legged locomotion.}
    Blind policies omit depth, perceptive policies consume depth, and
    binary-switching systems gate between perceptive and blind branches.
    \method{} instead keeps denoising world-model and proprioceptive
    pathways co-active and continuously fused. Training-only critics and
    privileged signals are omitted.}
  \label{fig:paradigm}
  \vspace{-8pt}
\end{figure}

Prior work approaches this spectrum from two ends and, recently, tries to
bridge them (Fig.~\ref{fig:paradigm}). \emph{Blind} policies infer
terrain- and body-state information from
proprioception~\citep{kumar2021rma, long2024him, nahrendra2023dreamwaq,
radosavovic2024nexttoken, zhao2024bilevel, liu2025locoformer} and tolerate
exteroceptive corruption by construction, but cannot anticipate obstacles
ahead. \emph{Perceptive} policies add
exteroception~\citep{zhuang2024humanoid, ben2025gallant,
sun2025perceptiveterrain, zhang2026rpl}, with some variants further
enriching the visual latent through reconstruction objectives such as
heightmap regression in PIE~\citep{luo2024pie} or world-model
autoencoding in WMP~\citep{lai2025wmp}. Yet they assume depth observations
remain clean and in-distribution, and degrade when this assumption fails.
\emph{Switching-based unification} systems combine a perceptive and a blind sub-policy and
route or switch between them based on perception
reliability~\citep{vbcom2026, liu2024mbc, zhang2025renet}. These
approaches leave a critical gap in the partial-corruption regime
between clean depth and complete loss: perceptive policies trust
degraded input as clean and degrade with it, while switching-based
systems bypass recoverable depth in favor of blind operation and can
introduce abrupt control changes. None recovers corrupted depth through
denoising.

We target this critical intermediate regime: partial corruption typically
leaves recoverable geometric cues in the depth history that prior policies
neither denoise nor exploit. We build this into \method{}, a single-stage
humanoid locomotion policy with two co-active pathways: a perceptive
world-model encoder that denoises corrupted depth, and a proprioceptive
variational encoder that supplies depth-free body-state information.
During training, the world-model decoder is supervised to reconstruct
clean depth from corrupted input. Unlike standard world-model autoencoding
of the observed input, this objective trains the perceptive latent to
recover clean depth structure from degraded observations. A coupled
training recipe pairs a depth-noise curriculum on the world-model input
with world-model feature dropout on the policy-facing latent, exercising
the policy across the full spectrum. \method{} thus adapts continuously
to perception quality: under partial corruption it recovers a denoised
depth estimate, and when forward-looking exteroception is no longer
informative it degrades toward proprioceptive operation without a
discrete mode switch.

In summary, our main contributions include:
\begin{itemize}
    \item \textbf{Continuously adaptive single-policy formulation.}
    We recast perception-robust humanoid locomotion as continuous adaptation
    rather than binary mode switching. \method{} continuously fuses a
    denoising world-model latent with a proprioceptive latent, recovering
    partially corrupted depth instead of immediately handing control to a
    blind branch.
    \item \textbf{Coupled degradation training recipe.}
    A depth-noise curriculum on the world-model input, paired with
    world-model feature dropout on the policy-facing latent, exposes the
    single policy to the full perception-quality spectrum and enables
    graceful degradation.
    \item \textbf{Simulation and real-world evaluation on the Unitree G1.}
    We evaluate graceful degradation through simulation perception-quality
    sweeps and real-robot tests spanning controlled occlusion,
    flash-induced sensor corruption, and outdoor depth artifacts.
\end{itemize}

\section{Related Work}
\label{sec:related}

\paragraph{Blind and Perceptive Legged Locomotion.}
Blind locomotion policies~\citep{kumar2021rma, long2024him,
nahrendra2023dreamwaq, wu2022daydreamer, cui2024adapting} walk
robustly on uneven terrain by reacting to contact forces, but their
reactive nature precludes anticipating gaps, high steps, and hurdles.
Perceptive policies therefore integrate exteroception, and existing
approaches largely fall into two categories. The
first relies on LiDAR to construct an explicit terrain
representation, typically a 2.5D elevation map, which is then fed to
the policy~\citep{long2025pim, beamdojo2025, ma2026cmoe}. Such
pipelines are sensitive to state estimation drift and LiDAR motion
distortion, breaking down under highly dynamic motion. The second
category foregoes explicit map construction and
feeds raw depth images directly into an end-to-end perceptive policy
on quadrupeds~\citep{zhuang2023robot, cheng2024extreme,
chane-sane2024soloparkour, li2025move} and
humanoids~\citep{zhuang2024humanoid, hiking2026inwild, more2025,
dpl2025, nowyousee2026}. Some methods further use auxiliary objectives,
such as heightmap reconstruction in PIE~\citep{luo2024pie} or
world-model autoencoding in WMP~\citep{lai2025wmp}, to enrich the
perceptive latent. Both categories
take the exteroceptive signal as given: clean,
complete, and in-distribution at deployment. Real-world exteroception
is unreliable: sensor noise, occlusion, motion-induced artifacts, and
outright signal loss are routine, and few existing methods examine
policy behaviour under sustained perception degradation. The recent
MGDP~\citep{dong2026mgdp} learns noise-resistant depth representations
via a contrastive U-Net trained to reconstruct clean targets from
corrupted inputs, but evaluates only moderate pixel-level corruption
and stays within the perceptive-policy regime, leaving graceful
operation across the full perception-quality spectrum unaddressed.

\paragraph{Unifying Perception and Proprioception.}
Several recent works combine perceptive and blind locomotion by
learning an arbitration mechanism between two sub-policies.
VB-Com~\citep{vbcom2026} gates a perceptive and a blind policy
with a learned return estimator;
MBC~\citep{liu2024mbc} co-trains perceptive and blind agents
under terrain-reconstruction regularization, with routing
learned implicitly rather than driven by an explicit detector;
RENet~\citep{zhang2025renet} runs vision-plus-proprio and
proprio-only estimators with a CNN-autoencoder selector,
switching to proprioception when the reconstruction error
exceeds a threshold. These designs differ in how the arbitration
signal is computed, but share the same failure-mode assumption:
unreliable perception should be bypassed by the blind branch.
This assumption is safe under complete sensor failure but treats
partially corrupted depth as if it were fully corrupted, where
the perception module could instead denoise its own latent.
Concurrent work~\citep{liu2026real} studies a complementary
axis, addressing visual degradation and blind zones through
Mamba~\citep{gu2024mamba}-based temporal memory and
physics-guided state estimation. This improves temporal
robustness, but does not train the perception module to map
partially corrupted depth into a latent from which the decoder
reconstructs clean depth. \method{} realizes this directly:
under partial corruption its latent supplies a denoised terrain
estimate to the actor, while a co-active proprioceptive
variational encoder, running at the control rate, supplies a
reliable body-state stream as corruption increases. The two
pathways fuse continuously, approaching blind-like operation
under severe perception failure rather than through a discrete
sub-policy handoff.

\section{Method}
\label{sec:method}

\afterpage{%
\begin{figure}[t]
  \centering
  \includegraphics[width=1.0\linewidth]{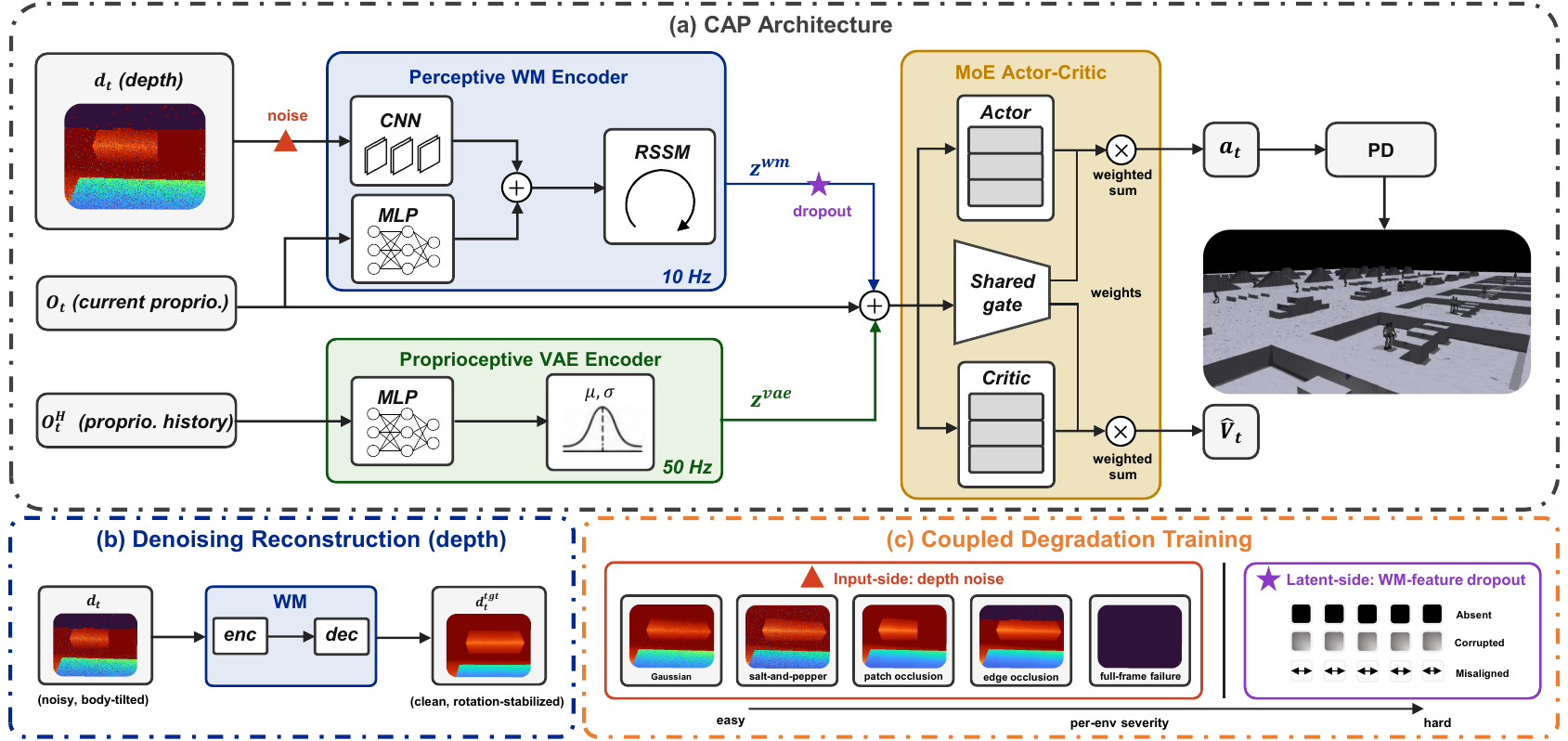}
  \caption{\textbf{\method{} framework.}
    (a) A perceptive world-model encoder and a co-active proprioceptive
    variational encoder feed a shared-gate mixture-of-experts
    actor--critic, with critic-only privileged signals omitted.
    (b) The world model reconstructs rotation-stabilized clean depth
    $d_t^{\mathrm{tgt}}$ from corrupted body-tilted depth $d_t$.
    (c) The same per-environment severity $n_e$ controls training-time
    input-side depth noise and latent-side world-model feature dropout
    on the policy-facing latent.}
  \label{fig:framework}
  \vspace{-8pt}
\end{figure}%
}

\subsection{Architecture Overview}
\label{sec:method-overview}

Fig.~\ref{fig:framework} summarizes \method{}'s architecture and
training recipe. Architecturally, \method{} couples an asymmetric
actor--critic with two co-trained encoders: a perceptive world-model
encoder and a proprioceptive variational encoder.
The asymmetric design~\citep{pinto2018asymmetric} lets us collapse
training into a single stage: the actor receives only signals
available on the deployed robot, whereas the critic additionally
accesses privileged signals. Optimisation has three concurrent
parts: PPO~\citep{schulman2017ppo} updates the actor and critic;
the variational autoencoder (VAE) is co-trained per PPO step under
its reconstruction--KL objective; and the world model (WM) is
updated by a separate optimiser alongside PPO.

\paragraph{Policy Network.}
The policy-side raw observations are a proprioceptive vector $o_t$
and an egocentric depth image $d_t$. The proprioceptive vector
\begin{equation}
  o_t = [\,\omega_t,\, g_t,\, c_t,\, \theta_t,\, \dot\theta_t,\,
            a_{t-1}\,]^{\!\top} \in \mathbb{R}^{45}
  \label{eq:actor-obs}
\end{equation}
stacks the body angular velocity $\omega_t$, projected gravity
$g_t$, velocity command $c_t$, joint positions $\theta_t$, joint
velocities $\dot\theta_t$, and the previous action $a_{t-1}$. The
depth image $d_t$ is captured from a forward-facing depth camera
at $10\,$Hz and may be corrupted at training time.

The two encoders compress these signals into latents:
a \emph{perceptive WM encoder} fuses $o_t$ and $d_t$ into
$z^{\text{wm}}_t$ at $10\,$Hz, while a
\emph{proprioceptive VAE encoder} compresses a proprioceptive
history $o_t^H = [o_{t-H+1}, \dots, o_t]$ of length $H$ into
$z^{\text{vae}}_t$ at $50\,$Hz. The actor input
concatenates the proprioception with the two latents,
\begin{equation}
  z^a_t = [\,o_t,\, z^{\text{wm}}_t,\, z^{\text{vae}}_t\,]^{\!\top},
  \label{eq:actor-input}
\end{equation}
and is routed through the actor branch of a shared-gate
mixture-of-experts (MoE), in which a gating network reads $z^a_t$
and combines the outputs of $K$ experts into the action $a_t$.
The gate produces a weighted combination of expert outputs from
the joint visual--proprioceptive representation, with both pathways
remaining co-active across perception conditions.

\paragraph{Value Network.}
The critic mirrors the actor's input and additionally accesses
privileged signals unavailable on the deployed robot:
\begin{equation}
  z^{c}_t = [\,z^{a}_t,\, v_t,\, e_t,\, m_t\,]^{\!\top},
  \label{eq:critic-input}
\end{equation}
where $v_t$ is the base linear velocity, $e_t$ is the external
disturbance force applied during domain randomization, and $m_t$
is a local heightmap scan sampled from terrain ground truth.
The heightmap provides a clean exteroceptive signal for value
estimation, independent of the actor-side depth path exposed to
corruption. The critic routes $z^{c}_t$ through $K$ experts that
reuse the actor's gating weights with a stop-gradient, preventing the
value loss from perturbing routing, and outputs $\hat{V}_t$.

\subsection{Dual-Pathway Encoders}
\label{sec:method-encoders}

The perception spectrum spans clean, partially corrupted, and
absent depth. These regimes place \emph{conflicting}
demands on the visual signal: under partial corruption the
policy must preserve and denoise useful terrain cues, while under
full perception failure it must operate as if blind. The temporal
memory that carries terrain structure through brief corruption can
become unreliable under sustained perception loss and
out-of-distribution (OOD) terrain. When depth is uninformative,
the WM predicts forward from learned dynamics fit to the training
terrain distribution; on unseen geometry, this prediction can
diverge from the true terrain geometry. We call this the WM's
\emph{overshoot} regime, which motivates both the dual-pathway
encoder design and WM-feature dropout.

To cover these regimes, the two encoders complement each other
along time-scale, modality, and failure mode. The WM encoder is the
$10\,$Hz \emph{learned-denoising} pathway: it uses depth and
proprioception to maintain a denoised, temporally predictive latent.
The VAE encoder is the $50\,$Hz \emph{high-rate proprioceptive}
pathway: it estimates short-horizon body state from proprioceptive
history at the control rate. Both pathways remain co-active and are
jointly consumed by the actor, rather than handed off between
regimes. Neither pathway alone is sufficient: the WM can drift in
the overshoot regime, while the VAE, without exteroceptive input,
cannot anticipate exteroception-required terrains such as
platforms, gaps, or hurdles.

\paragraph{Perceptive WM Encoder.}
We design the WM encoder to actively denoise the corrupted
depth signal. Unlike the standard Dreamer-style autoencoding
setup, in which the decoder reconstructs the encoder's own
input~\citep{lai2025wmp,hafner2025dreamerv3}, we
\emph{decouple input from target}: the encoder ingests the
observed depth $d_t$, while the decoder is trained
against a separate \emph{rotation-stabilized clean depth}
$d_t^{\mathrm{tgt}}$. The target view is rendered with the
camera roll and pitch fixed to nominal values (yaw still
follows body heading), so the objective forces the encoder
to learn (i)~a denoising map from corrupted to clean depth,
and (ii)~a transformation from the body-tilted sensor frame
into a horizon-aligned reference frame. Neither transformation is induced
by a standard autoencoding objective.

We instantiate the encoder as a DreamerV3 recurrent
state-space model (RSSM)~\citep{hafner2025dreamerv3} with
state $s_t = (h_t, z_t)$ comprising a deterministic recurrent
component $h_t$ and a stochastic latent $z_t$:
\begin{align}
  h_t                  &= f_\phi(h_{t-k},\, z_{t-k},\, a_{t-k:t-1})
                          & &\text{(recurrent)}, \notag\\
  z_t                  &\sim q_\phi(\cdot \mid h_t, o_t, d_t)
                          & &\text{(posterior)}, \notag\\
  \hat z_t             &\sim p_\phi(\cdot \mid h_t)
                          & &\text{(prior)}, \notag\\
  (\hat o_t, \hat d_t^{\mathrm{tgt}})
                       &\sim p_\phi(\cdot \mid h_t, z_t)
                          & &\text{(decoder)},
  \label{eq:rssm}
\end{align}
where $k$ is the number of policy steps between consecutive
WM updates, $o_t$ the proprioceptive observation, $d_t$ the
observed depth, and $a_{t-k:t-1}$ the sequence of
policy actions taken in between. The RSSM is
trained jointly with the policy by minimising
\begin{equation}
  \mathcal{L}_{\mathrm{WM}} = \mathbb{E}_{q_\phi}\!\Big[
    \underbrace{-\log p_\phi\!\left(d_t^{\mathrm{tgt}} \mid h_t, z_t\right)}_{\text{denoising recon}}
    - \underbrace{\log p_\phi\!\left(o_t \mid h_t, z_t\right)}_{\text{proprio recon}}
    + \beta_{\text{wm}}\, \mathrm{KL}\!\left( q_\phi(\cdot \mid h_t, o_t, d_t)\,\|\,p_\phi(\cdot \mid h_t) \right)
  \Big],
  \label{eq:wm-loss}
\end{equation}
in which the input--target mismatch $d_t \neq d_t^{\mathrm{tgt}}$
distinguishes $\mathcal{L}_{\mathrm{WM}}$ from prior
Dreamer-based formulations~\citep{lai2025wmp,hafner2025dreamerv3}.
The relative weighting of the reconstruction terms and the KL term in
$\mathcal{L}_{\mathrm{WM}}$ is balanced by a harmony-style
adaptive scaling~\citep{ma2024harmonydream}.
The prior $p_\phi(\cdot \mid h_t)$ lets the WM predict forward
when observations become unreliable. This temporal carry-over
helps bridge brief corruption, but can overshoot under sustained
perception loss.

A small MLP, learned with the actor under PPO, projects
$h_t$ into the actor-facing latent
$z^{\text{wm}}_t = \mathrm{MLP}_\psi(h_t)$ that enters
$z^a_t$, while the stochastic component $z_t$ stays internal to
the WM.

\paragraph{Proprioceptive VAE Encoder.}
The VAE encoder supplies a high-rate body-state latent from
proprioception. Running at the $50\,$Hz control rate and taking no
depth input, it complements the WM along modality and time-scale:
it tracks short-horizon body-state dynamics from proprioception,
while the WM integrates depth and proprioception at $10\,$Hz to
capture longer-horizon terrain structure. We instantiate it as a
small $\beta$-VAE~\citep{higgins2017betavae,nahrendra2023dreamwaq}
with parameters $\theta$, mapping the proprioceptive history $o_t^H$
to the body-state latent $z^{\text{vae}}_t$ and predicting the
next-step proprioception $\hat o_{t+1}$. The VAE is co-optimised
on every PPO iteration under
\begin{equation}
  \mathcal{L}_{\mathrm{VAE}} =
    \mathrm{MSE}\!\left(\hat o_{t+1},\, o_{t+1}\right)
    + \beta_{\text{vae}}\, \mathrm{KL}\!\left( q_\theta(z^{\text{vae}}_t \mid o_t^H)\,\|\,p(z^{\text{vae}}_t) \right).
  \label{eq:vae-loss}
\end{equation}

\subsection{Coupled Training for Graceful Degradation}
\label{sec:method-noise}

\paragraph{Depth Noise Curriculum.}
Real-world depth carries artifacts the simulator does not model.
We close this gap by injecting five composable corruption channels
into $d_t$ at training time, ordered from pixel-level to
frame-level disturbance:
\emph{Gaussian sensor noise} for read noise,
\emph{salt-and-pepper noise} for sporadic invalid pixels,
\emph{patch occlusion} for lens contamination,
\emph{edge occlusion} for parallax and self-occlusion at depth
discontinuities, and
\emph{full-frame failure} for sensor failures.
A per-env severity $n_e$ scales all five channels jointly, and
three profiles \textbf{Nominal\,/\,Degraded\,/\,Severe} correspond
to clean, partial perception loss, and complete
perception loss (channel-wise parameter ranges in
Appendix~\ref{app:depth_curriculum}). Within each profile, an adaptive update
tracks each env's rolling success rate $\bar s_e$ and steps
difficulty
\begin{equation}
  n_e \leftarrow
  \begin{cases}
    n_e + 1 & \bar s_e \geq \tau_{\uparrow}, \\
    n_e - 1 & \bar s_e \leq \tau_{\downarrow}, \\
    n_e     & \text{otherwise},
  \end{cases}
  \label{eq:noise-curr}
\end{equation}
clipped to $[0, N_{\max}]$, so difficulty is matched per env
automatically as the policy improves. On exteroception-required
terrains (e.g., Platform, Gap, and Hurdle), we exclude the
\emph{Severe} profile to keep the depth signal usable.

\paragraph{World-Model Feature Dropout.}
During training the WM produces a reliable $z^{\text{wm}}_t$.
Without regularization, the policy may rely on this signal heavily,
leaving the proprioceptive pathway under-exercised.
We address this with per-env WM-feature dropout. Its drop probability
$p_e = (n_e / N_{\max})\, p_{\max}$ shares the severity $n_e$
with the depth-noise curriculum, so each env trains under matched
input-side and latent-side corruption. When triggered, the
$z^{\text{wm}}_t$ entering the policy network is replaced with
one of three corruptions chosen uniformly:
\textbf{Absent} (zeroed out, removing WM support),
\textbf{Corrupted} (additive Gaussian noise at the per-feature
scale), and
\textbf{Misaligned} (cross-env shuffle that swaps in another
env's $z^{\text{wm}}_t$).

The two mechanisms cover complementary axes of perception
failure. Depth noise applies input-side corruption for
in-distribution WM denoising; WM-feature dropout applies
latent-side corruption that exposes the policy to OOD-like WM
prediction errors.

\section{Experiments}
\label{sec:exp}

\subsection{Experimental Setup}
\label{sec:setup}

All simulation experiments use the Unitree G1 humanoid in Isaac
Gym~\citep{makoviychuk2021isaac} with a common reward design on Stair,
Platform, Gap, and Hurdle. We evaluate two anchor cells,
\textbf{Clean} and \textbf{Noisy}, and an eight-stage perception-quality
sweep from nominal depth to severe failure, with the severe endpoint
tested only on Stair. Each Noisy or degraded stage fixes the severity
level, but its depth corruptions remain time-structured within a
rollout through per-frame noise, occlusion, and intermittent full-frame
failure. We report \textbf{Success Rate} (SR), the fraction of trials
reaching the goal, and \textbf{Avg.\ Ret.}

Fig.~\ref{fig:paradigm} defines the five paradigms. We compare
\method{} with \textbf{Hiking}~\citep{hiking2026inwild}, a perceptive
baseline without reconstruction; \textbf{PIE}~\citep{luo2024pie}, a
reconstruction-based perceptive baseline; and \textbf{Binary-switch}, a
RENet-inspired hard-switching variant~\citep{zhang2025renet}; Blind is
the perception-independent floor. Hiking and PIE use our depth-noise
curriculum, while the primary Binary-switch baseline is trained with
nominal sensor noise. We additionally evaluate a variant with matched
capacity and corruption training. We report
ablations where `$-$' denotes removing the corresponding component: the
shared-gate MoE (\textbf{$-$MoE}), depth-noise curriculum
(\textbf{$-$noise curr.}), or proprioceptive VAE encoder
(\textbf{$-$VAE encoder}). Auxiliary ablations and implementation details
appear in Appendices~\ref{app:implementation} and~\ref{app:simulation}.

\subsection{Anchor-Cell Comparison and Ablations}
\label{sec:anchor}

\begin{wraptable}{r}{0.5\linewidth}
  \vspace{-27pt}
  \centering
  \scriptsize
  \vspace{-\baselineskip}
  \caption{\textbf{Anchor-cell paradigm comparison and ablations under
    Clean and Noisy perception.} Terrain columns report Success
    Rate (\%)$\uparrow$, and the final columns report Avg.\ SR and Avg.\ Ret.
    }
  \label{tab:main}
  \setlength{\tabcolsep}{2.5pt}
\renewcommand{\arraystretch}{1.00}
\begin{tabular}{lcccccc}
\toprule
 & \multicolumn{4}{c}{Per-terrain SR (\%) $\uparrow$} & & \\
\cmidrule(lr){2-5}
Method & Stair & Plat. & Gap & Hurd. & Avg.\ SR $\uparrow$ & Avg.\ Ret.\ $\uparrow$ \\
\midrule
\rowcolor{black!8}\multicolumn{7}{l}{\textit{Clean}} \\
Hiking          & 92.3 & 93.7 & 96.3 & 83.0 & 91.3 & 38.3 \\
PIE             & 98.7 & 95.8 & 99.8 & 97.4 & 97.9 & 38.8 \\
Binary-switch   & 94.4 & 96.1 & 91.0 & 92.7 & 93.6 & 38.5 \\
\cmidrule(l){1-7}
$-$noise curr.  & \underline{98.9} & \underline{99.6} & \underline{100.0} & \textbf{99.8} & \underline{99.6} & \underline{41.7} \\
$-$VAE encoder  & 95.7 & \underline{99.6} & 99.9 & 93.4 & 97.1 & 40.4 \\
$-$MoE          & 87.3 & 98.0 & \underline{100.0} & 99.0 & 96.1 & 39.7 \\
\cmidrule(l){1-7}
\textbf{Ours}   & \textbf{99.3} & \textbf{99.9} & \textbf{100.0} & \underline{99.4} & \textbf{99.6} & \textbf{41.7} \\
\midrule
\rowcolor{black!8}\multicolumn{7}{l}{\textit{Noisy}} \\
Hiking          & 70.0 & 45.4 & 1.7  & 73.4 & 47.6 & 22.7 \\
PIE             & 95.7 & 84.3 & 96.7 & 76.8 & 88.4 & 35.2 \\
Binary-switch   & 80.6 & 0.0  & 0.0  & 0.0  & 20.1 & 14.5 \\
\cmidrule(l){1-7}
$-$noise curr.  & \underline{96.4} & 85.4 & 89.8 & 85.1 & 89.2 & 37.3 \\
$-$VAE encoder  & 93.8 & \underline{95.3} & 98.5 & 89.7 & \underline{94.3} & \underline{39.3} \\
$-$MoE          & 78.5 & 93.9 & \underline{99.1} & \underline{93.4} & 91.2 & 38.5 \\
\cmidrule(l){1-7}
\textbf{Ours}   & \textbf{98.0} & \textbf{97.2} & \textbf{99.1} & \textbf{97.2} & \textbf{97.9} & \textbf{40.6} \\
\bottomrule
\end{tabular}

  \vspace{-12pt}
\end{wraptable}

Table~\ref{tab:main} evaluates the two anchor cells: clean
depth and heavily corrupted but still partially informative
depth. The table should be read as two tests: whether robust
training compromises clean-depth locomotion, and whether a
method remains usable when depth is degraded.
\method{} passes both tests, achieving the best Avg.\ SR and
Avg.\ Ret.\ at both anchors ($99.6/97.9\%$ Avg.\ SR across
Clean/Noisy). Hiking collapses under corrupted depth despite
training with the same depth-noise curriculum; PIE is the strongest
non-WM perceptive baseline but still loses robustness under
corrupted depth; and Binary-switch retains usable Stair
performance at the Noisy anchor, where proprioception alone
can support progress, but loses the exteroception-required terrains
entirely.

The ablation rows separate the component roles. The
\textbf{$-$MoE} ablation reduces both Clean and Noisy
performance, with the largest Clean-cell loss on Stair. These
results support the performance benefit of the MoE architecture
under both clean and corrupted depth. The \textbf{$-$noise curr.}
ablation preserves Clean performance but loses robustness at
Noisy, showing that corrupted-depth robustness depends on the
depth-noise curriculum. The \textbf{$-$VAE encoder} ablation causes a consistent
loss at both anchors, especially on Stair and Hurdle,
showing that the high-frequency proprioceptive pathway
complements the lower-rate WM as a co-active stream rather
than a vision-absent fallback. The \textbf{$-$WM-feature dropout}
ablation causes only small in-distribution changes, so we keep
its evidence in Appendix~\ref{app:ablations} and examine the
OOD setting it targets in the case study of
Appendix~\ref{app:ood_dropout}.

\subsection{Perception-Quality Sweep}
\label{sec:perception_sweep}

\begin{wrapfigure}{r}{0.5\linewidth}
  \centering
  \small
  \vspace{-35pt}
  \includegraphics[width=\linewidth]{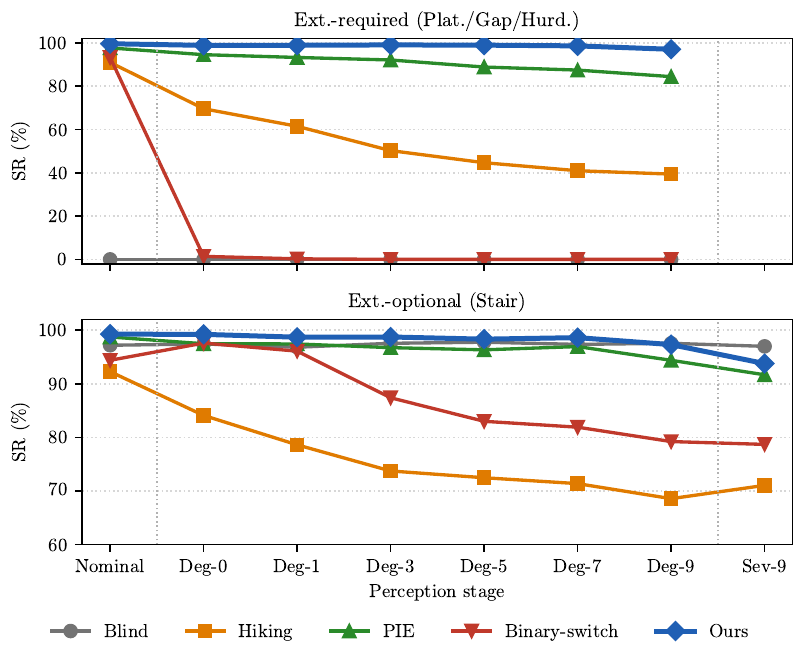}
  \caption{\textbf{Graceful degradation across the eight-stage sweep.}
    Top: exteroception-required average SR (Plat./Gap/Hurd.).
    Bottom: exteroception-optional Stair SR.}
  \label{fig:degradation}
  \vspace{-12pt}
\end{wrapfigure}

Fig.~\ref{fig:degradation} sweeps perception quality through
eight stages. \method{} turns gradual perception degradation into gradual
locomotion degradation, providing a continuous alternative to
discrete perception-blind switching.

On exteroception-required terrains, Blind stays near the
proprioceptive floor. \method{} degrades smoothly across the
Degraded sweep, while Hiking steadily loses performance without a
reconstruction objective, PIE declines more mildly with reconstruction
regularization, and Binary-switch suffers an abrupt drop when
its gate selects the blind branch. For Stair, where exteroception is
optional, the desired behavior is to approach the blind reference as
the exteroceptive signal becomes uninformative. \method{}
stays close to this reference through the Degraded sweep and
Severe tail, whereas Binary-switch remains below the blind
reference even under Severe failure, showing that hard branch
selection remains brittle when blind locomotion is viable.
This curve shape is the central behavioral evidence for graceful
degradation: smoothness across the perception-quality
spectrum is not generic to perceptive policies, but is achieved
by \method{}'s continuous fusion.

\subsection{Real-World Experiments}
\label{sec:real}

\begin{wraptable}{r}{0.45\linewidth}
  \vspace{-27pt}
  \centering
  \scriptsize
  \vspace{-\baselineskip}
  \caption{\textbf{Real-world G1 success rates across terrains and
    perception conditions} ($n{=}5$ trials per cell).}
  \label{tab:real}
  \setlength{\tabcolsep}{4pt}
\renewcommand{\arraystretch}{1.05}
\begin{tabular}{lccccc}
\toprule
 & \multicolumn{4}{c}{Per-terrain SR ($n=5$)} & \\
\cmidrule(lr){2-5}
Perception & Stair & Plat. & Gap & Mixed & Total \\
\midrule
Clean              & 5/5 & 5/5 & 5/5 & 5/5 & 20/20 \\
Partial occlusion  & 5/5 & 4/5 & 5/5 & 5/5 & 19/20 \\
Full cover         & 5/5 & 0/5 & 0/5 & 0/5  & 5/20  \\
\bottomrule
\end{tabular}

  \vspace{-12pt}
\end{wraptable}

The real-robot experiments test whether the continuous adaptation
observed in simulation preserves G1 locomotion under physical perception
degradation. We deploy the policy trained in simulation directly on the
Unitree G1 across controlled indoor perception perturbations and outdoor
scenes. Table~\ref{tab:real} reports the indoor trials across perception
conditions. Under Clean and Partial-occlusion settings, \method{}
completes 39/40 trials across Stair, Platform, Gap, and Mixed terrain.
Under Full cover, it succeeds in 5/5 Stair trials but 0/5 trials on each
of Platform, Gap, and Mixed. Full camera cover removes the visual cues
needed to anticipate platform edges and gaps. Proprioception alone is
insufficient to traverse these obstacles in the tested conditions.

\begin{figure}[t]
  \centering
  \includegraphics[width=\linewidth]{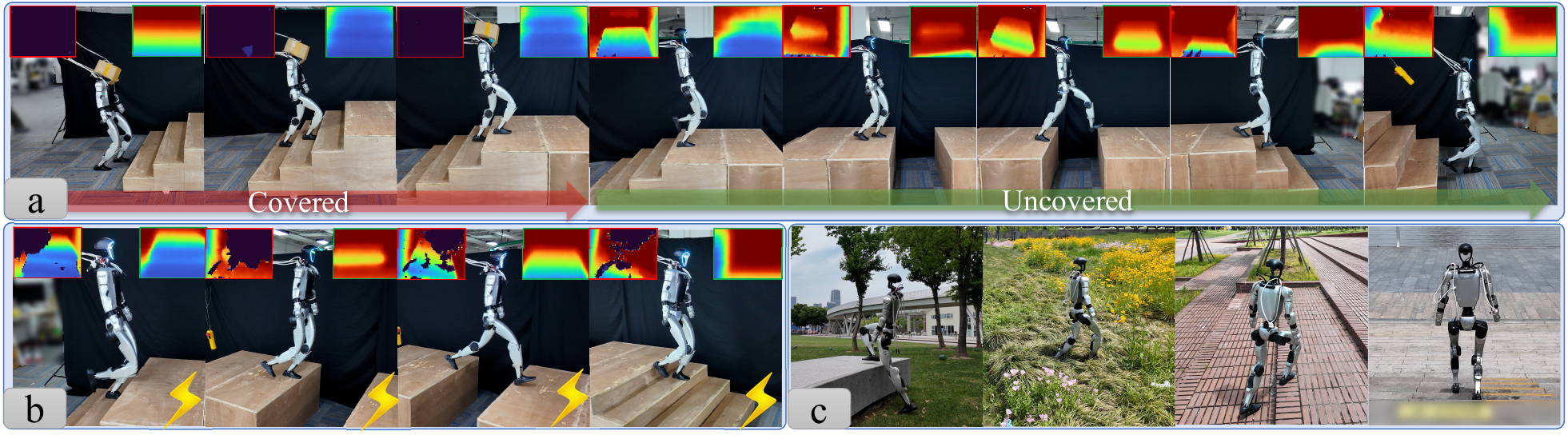}
  \caption{\textbf{Real-world deployment of \method{} on G1.}
    (a) Mixed-terrain cover--uncover. (b) Flash-induced depth
    corruption on mixed terrain. (c) Outdoor deployment on platform,
    grass, and stair scenes. Insets show synchronized raw depth
    inputs (left, red) and WM reconstructions (right, green).}
  \label{fig:real}
  \vspace{-8pt}
\end{figure}

Fig.~\ref{fig:real}(a) is the key transition case, with additional frames
in Fig.~\ref{fig:teaser}(a). In one mixed-terrain episode, the robot
ascends stairs under camera cover, then continues through the gap and
stair descent after perception is restored. This sequence tests
continuity through transient exteroceptive loss and re-acquisition.
Denoising is illustrated by Fig.~\ref{fig:teaser}(b): corrupted but
informative raw depth is paired with WM reconstructions that preserve
traversable structure, and the Partial-occlusion row of
Table~\ref{tab:real} provides the controlled quantitative counterpart.

Beyond physical occlusion, Fig.~\ref{fig:real}(b,c) tests the policy
under real-sensor artifacts and outdoor deployment. Flash corruption
creates large invalid regions in raw depth, while the WM reconstruction
preserves the main traversable structure. Outdoor platform, grass, and
stair trials expose the policy to vegetation-induced depth artifacts
beyond the simulator's noise channels. Together, the controlled success
rates and paired raw-depth and WM-reconstruction insets show
perception-robust hardware deployment under intermittent occlusion,
real-sensor corruption, and outdoor depth artifacts.

\section{Conclusion, Limitations and Future Directions}
\label{sec:conc}

\paragraph{Conclusion.}
We presented \method{}, a single-stage policy that adapts continuously
to perception quality through a denoising world-model encoder, a
co-active proprioceptive variational encoder, and input- and
feature-level degradation training. Simulation and Unitree G1
experiments show that this design turns partial perception failure into
graceful degradation within one policy, rather than a discrete
perceptive-to-blind handoff.

\paragraph{Limitations and Future Directions.}
\method{} does not yet address safety at the boundary of
perception-conditioned locomotion. Large out-of-distribution geometric
errors (e.g., substantial camera-extrinsic drift) may still make the denoised percept
unreliable, and under complete perception loss on exteroception-required
terrain, degrading toward blind locomotion must be paired with recovery
from failed steps or fall-prone states. Future work should extend
\method{} toward locomotion safety through reliability estimation,
risk-aware action selection, and safe recovery.

\section*{Acknowledgments}
This work was supported by the National Natural Science Foundation of China (NSFC) under Grant 62403142.

\bibliography{refs}

\clearpage
\appendix
\setcounter{page}{1}
\renewcommand{\thepage}{A\arabic{page}}
\setcounter{table}{0}
\setcounter{figure}{0}
\renewcommand{\thetable}{A\arabic{table}}
\renewcommand{\thefigure}{A\arabic{figure}}
\nolinenumbers %
\section*{Appendix Overview}

The appendix provides supporting details for the method, simulation
experiments, and real-world deployment. Appendix~\ref{app:implementation}
gives implementation details for \method{}, including network
architecture, the perceptive WM encoder interface, reward terms,
domain randomization, and depth-camera randomization.
Appendix~\ref{app:simulation} describes the
simulation setup, baseline recipes, auxiliary ablations, and a
full-cover stair-to-plane recovery case study for WM-feature dropout.
Appendix~\ref{app:real_details} describes the real-world deployment
setup and additional deployment examples.

\section{Implementation Details}
\label{app:implementation}

\subsection{Network Architecture}
\label{app:arch}

The actor consumes the current proprioceptive observation
$o_t\in\mathbb{R}^{45}$ together with the WM and VAE latents,
$z^{\text{wm}}_t\in\mathbb{R}^{32}$ and
$z^{\text{vae}}_t\in\mathbb{R}^{16}$. The concatenated
$93$-dimensional vector is routed by a shared-gate MoE with
$K{=}4$ actor experts. The critic augments the same actor-side
representation with the privileged base velocity $v_t$, external
disturbance $e_t$, and local heightmap $m_t$, and reuses the actor
gate with stop-gradient for value estimation.

The WM encoder runs with the $10\,\mathrm{Hz}$ depth stream and
projects the deterministic RSSM state to $z^{\text{wm}}_t$; the
VAE encoder runs at the $50\,\mathrm{Hz}$ control rate from a
five-step proprioceptive history. Table~\ref{tab:app_network}
lists the per-module network dimensions, and
Table~\ref{tab:app_training} lists the training and runtime
hyperparameters.

\begin{table}[!htbp]
  \centering
  \caption{Per-module network architecture of \method{}. Hidden
  sizes are listed from input to output; ``$\to$'' denotes the
  layer that produces the listed output. All modules use ELU
  activations.}
  \label{tab:app_network}
  \footnotesize
  \setlength{\tabcolsep}{4pt}
  \renewcommand{\arraystretch}{1.12}
  \begin{tabular}{@{}
    >{\raggedright\arraybackslash}p{0.14\linewidth}
    >{\raggedright\arraybackslash}p{0.31\linewidth}
    >{\raggedright\arraybackslash}p{0.49\linewidth}
  @{}}
    \toprule
    Component & Item & Value \\
    \midrule
    \multicolumn{3}{@{}l}{\textit{MoE actor}} \\
    & Current proprioception $o_t$ & $45$ \\
    & WM latent $z^{\text{wm}}_t$ / VAE latent $z^{\text{vae}}_t$ & $32$ / $16$ \\
    & Concatenated actor input & $45{+}32{+}16=93$ \\
    & Shared gate & $128\to$ softmax over $K{=}4$ experts \\
    & Expert MLP & $[256,128,64]$ \\
    & Action head & $\to12$, diagonal Gaussian \\
    \midrule
    \multicolumn{3}{@{}l}{\textit{Critic}} \\
    & Input & actor repr. $+\,v_t\,(3),\,e_t\,(3),\,m_t\,(187)$ \\
    & Privileged encoder & $[256,128]\to64$ \\
    & Heightmap encoder & $[64,64]\to32$ \\
    & WM encoder & $[64,64]\to32$ \\
    & Expert MLP & $[512,256,128]$ \\
    & Gate & shared actor gate (stop-gradient) \\
    & Value head & $\to1$ \\
    \midrule
    \multicolumn{3}{@{}l}{\textit{WM encoder (DreamerV3 RSSM)}} \\
    & Depth input & $64\times48$ at $10\,\mathrm{Hz}$ \\
    & Deterministic state & $512$ \\
    & Stochastic state & $32\times32$ (categorical) \\
    & Policy-facing projection & $[64,64]\to32$ \\
    \midrule
    \multicolumn{3}{@{}l}{\textit{VAE encoder}} \\
    & Proprioceptive history & $H{=}5$ steps at $50\,\mathrm{Hz}$ \\
    & Encoder / decoder & $[128,32]$ / $[32,128]$ \\
    & Latent $z^{\text{vae}}_t$ & $16$ \\
    \bottomrule
  \end{tabular}
\end{table}

\begin{table}[!htbp]
  \centering
  \caption{Training and runtime hyperparameters for \method{}.
  PPO, the world model, and the VAE encoder are optimised jointly,
  each with its own optimiser; PPO uses a KL-adaptive learning rate,
  while the world-model and VAE learning rates are held constant.}
  \label{tab:app_training}
  \footnotesize
  \setlength{\tabcolsep}{4pt}
  \renewcommand{\arraystretch}{1.14}
  \begin{tabular}{@{}
    >{\raggedright\arraybackslash}p{0.18\linewidth}
    >{\raggedright\arraybackslash}p{0.45\linewidth}
    >{\raggedright\arraybackslash}p{0.31\linewidth}
  @{}}
    \toprule
    Component & Hyperparameter & Value \\
    \midrule
    \multicolumn{3}{@{}l}{\textit{Simulation \& runtime}} \\
    & Parallel environments & $16384$ \\
    & Simulation step & $5\,\mathrm{ms}$ \\
    & PD control frequency & $200\,\mathrm{Hz}$ \\
    & Policy frequency (decimation $4$) & $50\,\mathrm{Hz}$ \\
    & WM frequency ($k{=}5$ policy steps) & $10\,\mathrm{Hz}$ \\
    & Episode length & $20\,\mathrm{s}$ \\
    \midrule
    \multicolumn{3}{@{}l}{\textit{PPO}} \\
    & Training iterations & $20{,}000$ \\
    & Rollout length per env & $24$ \\
    & Learning epochs / minibatches & $5$ / $4$ \\
    & Optimiser & Adam \\
    & Learning rate & $10^{-3}$, KL-adaptive \\
    & Target KL divergence & $0.01$ \\
    & Clip range & $0.2$ \\
    & Discount $\gamma$ / GAE $\lambda$ & $0.99$ / $0.95$ \\
    & Entropy / value-loss coef. & $0.01$ / $1.0$ \\
    & Max gradient norm & $1.0$ \\
    \midrule
    \multicolumn{3}{@{}l}{\textit{World model}} \\
    & Optimiser & Adam \\
    & Learning rate & $10^{-4}$ \\
    & Batch size / sequence length & $16$ / $48$ \\
    & Gradient steps per PPO iter. & $10$ \\
    \midrule
    \multicolumn{3}{@{}l}{\textit{VAE encoder}} \\
    & Optimiser & Adam \\
    & Learning rate & $10^{-3}$ \\
    & KL weight $\beta_{\text{vae}}$ & $0.005$ \\
    & Max gradient norm & $10$ \\
    \bottomrule
  \end{tabular}
\end{table}

\subsection{Perceptive WM Encoder Interface}
\label{app:rssm}

This subsection details the RSSM-based implementation of the
perceptive WM encoder introduced in \S\ref{sec:method-encoders};
it specifies how the encoder is integrated into the control loop
rather than restating RSSM background~\citep{hafner2025dreamerv3}.
The concrete state sizes and update rates are listed in
Tables~\ref{tab:app_network} and~\ref{tab:app_training}.
The encoder is used purely as a learned-denoising perception
pathway, not as an imagination engine: the actor and critic are
trained by PPO on transitions collected in the simulator, and the
world model never generates imagined rollouts for policy
optimisation. Accordingly, only the projection
$z^{\text{wm}}_t = \mathrm{MLP}_\psi(h_t)$ of the deterministic
state $h_t$ is exposed to the actor; the stochastic state $z_t$ and
the decoder are used only to train the model.

\paragraph{Encoder, decoder, and depth target.}
The posterior $q_\phi(\cdot \mid h_t, o_t, d_t)$ fuses two
observation streams through modality-specific encoders---depth
$d_t$ through a convolutional network and proprioception $o_t$
through an MLP---and a multi-head decoder reconstructs depth and
proprioception from $(h_t, z_t)$ with a convolutional head and an
MLP head, matching the two reconstruction terms of
$\mathcal{L}_{\mathrm{WM}}$ (Eq.~\eqref{eq:wm-loss}). The depth
input and target are decoupled by rendering two images per timestep
from the same camera position and yaw heading: the encoder ingests
the corrupted, body-tilted view $d_t$, while the target
$d_t^{\mathrm{tgt}}$ is the clean simulator depth at the same
instant, rendered with the camera roll and pitch reset to nominal
and the yaw still tracking the body heading.

\paragraph{Low-frequency update and prior rollout.}
The RSSM state is refreshed from visual observations once every
$k{=}5$ policy steps, giving the $10\,\mathrm{Hz}$ world-model rate
against the $50\,\mathrm{Hz}$ control rate. On a refresh step a
posterior step ingests the current depth and proprioception and
updates the state; between refreshes, the actor reuses the cached
latent $z^{\text{wm}}_t$ while the state is advanced open-loop by
prior steps. The cache/prior path bridges brief perception loss,
and is the source of the \emph{overshoot} regime under sustained
loss on out-of-distribution terrain.

\subsection{Reward Functions and Domain Randomization}
\label{app:reward}

All policies use the same policy reward terms and domain-randomization
settings unless an ablation explicitly removes a training component.
Table~\ref{tab:app_reward} lists the reward terms used for policy
training, and Table~\ref{tab:app_dr} lists the robot-side
randomization ranges. Camera pose, camera intrinsics, and depth-image
corruption are shared across methods and reported separately in
Appendix~\ref{app:depth_curriculum}.

\begin{table}[!htbp]
  \centering
  \caption{Reward terms used for G1 parkour-policy training. Positive weights
  denote task rewards and negative weights denote penalties.
  $v$, $\omega$, $g$, $q$, $\tau$, and $a$ denote base linear
  velocity, base angular velocity, projected gravity, joint
  positions, joint torques, and the action; a $^{\mathrm{cmd}}$
  superscript marks a commanded quantity.}
  \label{tab:app_reward}
  \footnotesize
  \setlength{\tabcolsep}{3pt}
  \renewcommand{\arraystretch}{1.2}
  \begin{tabular}{@{}p{0.30\linewidth}p{0.44\linewidth}p{0.15\linewidth}@{}}
    \toprule
    Reward term & Definition & Weight \\
    \midrule
    \multicolumn{3}{@{}l}{\textit{Task reward}} \\
    Global velocity tracking & $\mathrm{clip}\!\big(v^{\mathrm{cmd}}-\Delta,\,0\big)/(|v^{\mathrm{cmd}}|+\varepsilon)$\,$^{\dagger}$ & $2.0$ \\
    Yaw tracking & $\exp\!\big(-|\mathrm{wrap}(\psi^{\mathrm{cmd}}-\psi)|\big)$ & $2.0$ \\
    Lateral velocity & $v_\perp^2$, velocity perpendicular to heading & $-2.0$ \\
    \midrule
    \multicolumn{3}{@{}l}{\textit{Base and posture regularization}} \\
    Vertical velocity & $v_z^2$ & $-1.0$ \\
    Roll/pitch angular velocity & $\omega_x^2+\omega_y^2$ & $-0.05$ \\
    Orientation & $g_x^2+g_y^2$ & $-2.0$ \\
    Base height & $(h-h^\star)^2$, $h^\star{=}0.72\,\mathrm{m}$ & $-12.0$ \\
    Hip posture & $\sum_{i\in\{\mathrm{hip\ roll,yaw}\}}(q_i-q_i^{\mathrm{def}})^2$ & $-0.5$ \\
    \midrule
    \multicolumn{3}{@{}l}{\textit{Foot-contact and safety terms}} \\
    Feet air time & $\sum_f (t^{\mathrm{air}}_f-0.5)\,\mathbb{I}[\text{first contact}]$ & $1.0$ \\
    Feet lateral distance & foot lateral spacing toward $[0.18,0.24]\,\mathrm{m}$ & $0.1$ \\
    Feet ground parallel & per-foot sample-height variance & $-0.02$ \\
    Feet stumble & $\mathbb{I}\big[\|F^{xy}_{\mathrm{feet}}\|^2 > 9\,|F^{z}_{\mathrm{feet}}|^2\big]$ & $-1.0$ \\
    Feet edge & foot contact on terrain edge cells & $-1.0$ \\
    Collision & contact on penalized body links & $-15.0$ \\
    \midrule
    \multicolumn{3}{@{}l}{\textit{Actuation and limit regularization}} \\
    DOF acceleration & $\sum_i \ddot q_i^2$ & $-2.5{\times}10^{-7}$ \\
    DOF velocity & $\sum_i \dot q_i^2$ & $-5.0{\times}10^{-4}$ \\
    Torques & $\sum_i \tau_i^2$ & $-1.0{\times}10^{-5}$ \\
    Action rate & $\|a_t-a_{t-1}\|_2^2$ & $-0.3$ \\
    Position limits & joint-position limit violation & $-2.0$ \\
    Velocity limits & joint-velocity limit violation & $-1.0$ \\
    Torque limits & torque-limit violation & $-1.0$ \\
    \bottomrule
  \end{tabular}

  \vspace{2pt}
  {\footnotesize $^{\dagger}$\,$\Delta$ is an asymmetric
  under/over-speed penalty on the heading-projected forward
  velocity.}
\end{table}

\begin{table}[!htbp]
  \centering
  \caption{Robot domain randomization used during training. Camera and
  depth-image randomization are reported in
  Appendix~\ref{app:depth_curriculum}.}
  \label{tab:app_dr}
  \footnotesize
  \setlength{\tabcolsep}{5pt}
  \renewcommand{\arraystretch}{1.2}
  \begin{tabular}{@{}p{0.42\linewidth}p{0.42\linewidth}@{}}
    \toprule
    Parameter & Range / value \\
    \midrule
    Torso payload mass & $[-5,5]\,\mathrm{kg}$ \\
    Body CoM displacement & $[-0.05,0.05]\,\mathrm{m}$ per axis \\
    Ground friction & $[0.0,1.25]$ \\
    Joint $K_p$ and $K_d$ multipliers & $[0.9,1.1]\times$ default \\
    Motor position offset & $[-0.01,0.01]\,\mathrm{rad}$ \\
    Initial joint-position scale & $[0.8,1.2]\times$ default \\
    Initial joint-position offset & $[-0.1,0.1]\,\mathrm{rad}$ \\
    Action delay & $0$--$3$ simulation substeps ($0$--$15\,\mathrm{ms}$) \\
    Robot push velocity & $[-0.75,0.75]\,\mathrm{m/s}$ per $xy$ axis every $16\,\mathrm{s}$ \\
    External disturbance force & $[-50,50]\,\mathrm{N}$ per axis every $8$ control steps \\
    \bottomrule
  \end{tabular}
\end{table}

\FloatBarrier

\subsection{Depth and Camera Randomization}
\label{app:depth_curriculum}

Training randomizes both the depth image observed by the WM encoder and
the depth-camera geometry. The five depth-corruption channels are shown
visually in Fig.~\ref{fig:framework} and parameterized in
Table~\ref{tab:app_depth_noise}; camera perturbations are listed in
Table~\ref{tab:app_camera_randomization}.

For the less standard channels, patch occlusion samples up to six
rectangular masks per depth frame; each active mask has a random
location and size and is filled with the near clip, far clip, or a
random depth value. Edge occlusion samples one image border and masks a
jittered strip whose thickness is controlled by the depth ratio; the
mask is either episode-persistent or transient for the listed number of
depth frames. Full-frame failure replaces the entire depth image by the
near clip, far clip, or a random constant during a contiguous interval
whose duration is determined by the time ratio.

The per-env severity $n_e\in[0,10]$ interpolates within the ranges in
Table~\ref{tab:app_depth_noise}, following the adaptive update in
Eq.~\eqref{eq:noise-curr}. On exteroception-required terrains
(Platform, Gap, and Hurdle), the
\textbf{Severe} profile is automatically demoted to \textbf{Degraded}
and the full-frame failure time ratio is capped at $0.05$ to keep the
depth signal usable.

\begin{table}[!htbp]
  \centering
  \caption{Depth-corruption parameter ranges for the three training
  severity profiles. Each env's severity $n_e \in [0,10]$
  interpolates within the listed bounds.}
  \label{tab:app_depth_noise}
  \footnotesize
  \setlength{\tabcolsep}{4pt}
  \renewcommand{\arraystretch}{1.1}
  \begin{tabular}{@{}
    >{\raggedright\arraybackslash}p{0.36\linewidth}
    >{\raggedright\arraybackslash}p{0.17\linewidth}
    >{\raggedright\arraybackslash}p{0.17\linewidth}
    >{\raggedright\arraybackslash}p{0.17\linewidth}
  @{}}
    \toprule
    Parameter & Nominal & Degraded & Severe \\
    \midrule
    \multicolumn{4}{@{}l}{\emph{Gaussian sensor noise}} \\
    Episode bias std.
      & $[0.005,0.025]$
      & $[0.05,0.4]$
      & $[0.05,0.4]$ \\
    Per-step std.
      & $[0.01,0.05]$
      & $[0.05,0.4]$
      & $[0.05,0.4]$ \\
    \midrule
    \multicolumn{4}{@{}l}{\emph{Salt-and-pepper invalid pixels}} \\
    Per-step invalid-pixel probability
      & $[0.01,0.02]$
      & $[0.1,0.6]$
      & $[0.1,0.6]$ \\
    \midrule
    \multicolumn{4}{@{}l}{\emph{Patch occlusion}} \\
    Block size (pixels)
      & $[10,15]$
      & $[20,30]$
      & $[20,30]$ \\
    Per-step block probability
      & $[0.05,0.1]$
      & $[0.15,0.3]$
      & $[0.15,0.3]$ \\
    \midrule
    \multicolumn{4}{@{}l}{\emph{Edge occlusion}} \\
    Activation probability
      & $[0.01,0.05]$
      & $[0.05,0.3]$
      & $[0.15,0.4]$ \\
    Occluded depth ratio
      & $[0.05,0.15]$
      & $[0.15,0.6]$
      & $[0.25,0.75]$ \\
    Persistence (depth frames)
      & $[5,15]$
      & $[15,40]$
      & $[20,50]$ \\
    \midrule
    \multicolumn{4}{@{}l}{\emph{Full-frame failure}} \\
    Episode time ratio
      & $0$
      & $[0.05,0.4]$
      & $[0.2,1.0]$ \\
    \bottomrule
  \end{tabular}
\end{table}

\begin{table}[!htbp]
  \centering
  \caption{Depth-camera randomization used during training. The same
  camera perturbation ranges are used for all severity profiles.}
  \label{tab:app_camera_randomization}
  \footnotesize
  \setlength{\tabcolsep}{5pt}
  \renewcommand{\arraystretch}{1.15}
  \begin{tabular}{@{}p{0.44\linewidth}p{0.42\linewidth}@{}}
    \toprule
    Parameter & Range / value \\
    \midrule
    Camera position offset & $\pm[0.02,0.02,0.02]\,\mathrm{m}$ \\
    Camera roll/pitch/yaw offset & $\pm[1^\circ,2^\circ,1^\circ]$ \\
    Focal-length scale, $f_x$ & $[0.9,1.15]\times$ nominal \\
    Focal-length scale, $f_y$ & $[0.85,1.1]\times$ nominal \\
    Principal-point offset, $c_x,c_y$ & $\pm0.05\times$ nominal image center \\
    \bottomrule
  \end{tabular}
\end{table}

\FloatBarrier

\section{Simulation Experiments}
\label{app:simulation}

\subsection{Simulation Setup}
\label{app:eval}

The simulation evaluation cells use fixed depth-corruption stages rather
than the adaptive training curriculum. Clean uses the \textbf{Nominal}
profile with $n_e \sim \mathrm{U}[0,3]$, Noisy uses the
\textbf{Degraded} profile with $n_e \sim \mathrm{U}[7,9]$, and Full
failure uses the \textbf{Severe} profile with $n_e{=}9$ on Stairs only.

We evaluate every method on $4096$ environments split equally across the
four reported simulation terrain categories (Stair, Platform, Gap, and
Hurdle) under three random seeds, and report the seed mean. Auto-reset is
disabled at evaluation, so each environment contributes a single trajectory
and its first termination locks the outcome; each environment's noise
realisation is fixed across methods by seed, isolating architecture and
training recipe from the evaluation distribution. All policies follow a
constant forward command of $0.8\,\mathrm{m/s}$ over a $20\,\mathrm{s}$
horizon. A trial
succeeds when the robot reaches the end-of-tile goal region before any
failure condition locks the episode; the goal is placed
$0.5\,\mathrm{m}$ before the end of the $10\,\mathrm{m}$ tile. Failure
conditions are time-out, lateral deviation beyond $0.75\,\mathrm{m}$,
posture collapse (base height below $0.3\,\mathrm{m}$ above the local
terrain, or roll or pitch beyond $1.0\,\mathrm{rad}$), or leaving the
evaluation envelope ($|d_x|>20\,\mathrm{m}$ or
$|d_y|>10\,\mathrm{m}$, where $d_x$ and $d_y$ are offsets from the tile
origin) before reaching the goal. Success Rate is the fraction of successful
environments in a cell, and Avg.\ Ret.\ is the mean episode return over
all environments in the cell, accumulated until each environment's
terminating event.

All terrains are generated on a $10\times10\,\mathrm{m}$ tile, with
difficulty $d\in[0,1]$ controlling the obstacle geometry. Evaluation
disables terrain progression and samples the high-difficulty band
$d\sim\mathrm{U}[0.8,1.0)$. Stair linearly increases step height from
$0.05$ to $0.23\,\mathrm{m}$ over the full difficulty range; Platform
increases step height from $0.10$ to $0.45\,\mathrm{m}$; Gap increases
gap width from $0.10$ to $0.85\,\mathrm{m}$; Hurdle increases obstacle
height up to $0.45\,\mathrm{m}$. The traversable corridor width also
narrows with difficulty.

\subsection{Baseline Implementation Details}
\label{app:baselines}

\textbf{Binary-switch} is a controlled adaptation of the
binary-switching paradigm rather than a verbatim reimplementation of
RENet~\citep{zhang2025renet}; we implement hard switching within
\method{}'s framework to compare binary branch selection with
continuous fusion.

Concretely, the actor always receives the current proprioceptive observation
and command, matching \method{}'s actor-side interface, and applies a binary
mask only to the two auxiliary slots: the WM latent and a proprioceptive
latent. The mask zeroes exactly one of these two slots.
At deployment the mask is driven by a low-pass-filtered WM reconstruction
loss (smoothing factor $0.1$): once the smoothed loss exceeds a threshold,
the policy switches from the perceptive slot to the proprioceptive slot. The
threshold is auto-calibrated as the $98$th percentile of the reconstruction
loss over a $500$-step clean rollout, after a short warm-up that lets the
recurrent state settle. During training the mask alternates between a
vision-primary and a proprioception-primary phase every $20$ PPO iterations,
with the proprioception-primary phase restricted to terrain classes that are
traversable without exteroception (plane, slope, stairs, discrete, and
parkour-stair tiles). The primary Binary-switch baseline is trained
with nominal sensor noise. It does not use the depth-noise curriculum,
WM-feature dropout, or heightmap-augmented critic.

\textbf{Hiking}~\citep{hiking2026inwild} follows its end-to-end perceptive
recipe: a convolutional depth encoder feeds a mixture-of-experts actor and
critic with no reconstruction objective. \textbf{PIE}~\citep{luo2024pie}
follows its native recipe, pairing a recurrent proprioceptive encoder and a
convolutional depth encoder with an autoencoding heightmap-reconstruction
objective. We reproduce both faithfully and, to match \method{}'s training
conditions, train the reported Hiking and PIE policies under our depth-noise
curriculum; the vanilla variants without the curriculum collapse once depth
corruption is introduced, so we report only the curriculum-matched variants
and add no separate vanilla result tables.

All baselines share \method{}'s reward function, terrain suite, and
evaluation protocol, so differences in success reflect architecture and
training recipe rather than the evaluation distribution.

\subsection{Matched Binary-switch Comparison}
\label{app:binary_matched}
We train an additional Binary-switch variant with the same policy and
world-model backbone, matched training capacity, and \method{}'s
depth-noise curriculum. The reward, terrain suite, and training budget
are held fixed. We select the switching threshold on a validation set,
then fix it for testing on separate seeds. Starting from the
clean-calibrated threshold $\beta_0$, we search multipliers
$\{0.25, 0.5, 0.75, 1, 1.25, 1.5, 2, 4, 8, 16\}$ using three
validation seeds at Nominal, Deg-3, and Deg-9. We choose the threshold
with the highest mean success rate across these conditions and terrains.

\begin{table}[!htbp]
  \centering
  \caption{\textbf{Matched Binary-switch comparison.} Success rate (\%)
  under a shared evaluation protocol, using three evaluation seeds and
  4096 environments per seed at each stage. Nominal, Low, and High report
  macro averages over Stair, Platform, Gap, Hurdle, and Mixed.
  Low and High average results at Degraded levels
  $[0,3]$ and $[7,9]$, respectively. Full cover reports Stair only.}
  \label{tab:binary_matched}
  \small
  \setlength{\tabcolsep}{7pt}
  \renewcommand{\arraystretch}{1.1}
  \begin{tabular}{lrrrr}
    \toprule
    Method & Nominal & Low corruption & High corruption & Full cover \\
    \midrule
    Binary-switch (matched) & 97.96 & 91.90 & 86.27 & 84.96 \\
    \method{}               & \textbf{99.40} & \textbf{98.68} & \textbf{97.59} & \textbf{93.78} \\
    \bottomrule
  \end{tabular}
\end{table}

Validation selects the largest threshold in the predefined grid,
$16\beta_0$. With this threshold fixed, the mean proprioception-only
branch duty remains below $1\%$ at every test stage.
Table~\ref{tab:binary_matched} shows that \method{} achieves higher
success rates across all four reported conditions.

These results suggest that corruption-aware training reduces the need
for a hard handoff to the proprioceptive branch. The perceptive branch
learns to operate with degraded depth, so partial corruption does not
necessarily make its features unusable. Switching away can therefore
discard useful terrain information. The preference for retaining
perceptive features supports \method{}'s design of keeping both pathways
co-active.

\subsection{Auxiliary Ablations}
\label{app:ablations}

Table~\ref{tab:app_aux_ablation} reports two secondary ablations at the
Clean and Noisy anchor cells: removing WM-feature dropout and removing the
heightmap-augmented critic. The \textbf{Ours} reference row matches
Table~\ref{tab:main}, and all averages are taken over the four obstacle
terrains.

These auxiliary rows are consistent with the component interpretation in
the main table. Removing WM-feature dropout causes only small
in-distribution changes: its SR remains close to \method{} at both anchors,
and Avg.\ Ret.\ is comparable or slightly higher in these cells. This
supports the view that WM-feature dropout has only a small
in-distribution effect; its role is to regularize the policy for OOD
WM errors, which we examine in Appendix~\ref{app:ood_dropout}.

Removing the heightmap-augmented critic has a more visible effect on both
metrics. Success drops on the exteroception-required terrains, especially
under Noisy perception, and Avg.\ Ret.\ decreases in both anchor cells.
This matches the design motivation in Section~\ref{sec:method}: the
privileged heightmap gives the critic a clean value-estimation signal that
is independent of the actor-side depth path exposed to corruption.

\begin{table}[!htbp]
  \centering
  \caption{\textbf{Auxiliary ablations at the Clean and Noisy anchor
  cells.} Per-terrain success rate (SR, \%) is reported for the four
  obstacle terrains, and Avg.\ SR / Avg.\ Return are averaged over these
  four terrains. The \textbf{Ours} reference row matches
  Table~\ref{tab:main}. Bold marks the best entry per column within a
  cell, underline the second best.}
  \label{tab:app_aux_ablation}
  \footnotesize
  \setlength{\tabcolsep}{6pt}
  \renewcommand{\arraystretch}{1.15}
  \begin{tabular}{lcccccc}
    \toprule
     & \multicolumn{4}{c}{Per-terrain SR (\%) $\uparrow$} & & \\
    \cmidrule(lr){2-5}
    Method & Stair & Platform & Gap & Hurdle & Avg.\ SR $\uparrow$ & Avg.\ Return $\uparrow$ \\
    \midrule
    \rowcolor{black!8}\multicolumn{7}{l}{\textit{Clean}} \\
    $-$\,heightmap-augmented critic & 96.8 & 96.7 & \underline{100.0} & 97.4 & 97.7 & 40.0 \\
    $-$\,WM-feature dropout         & \underline{99.1} & \underline{99.6} & \underline{100.0} & \underline{98.9} & \underline{99.4} & \textbf{42.0} \\
    \cmidrule(l){1-7}
    \textbf{Ours}                   & \textbf{99.3} & \textbf{99.9} & \textbf{100.0} & \textbf{99.4} & \textbf{99.6} & \underline{41.7} \\
    \midrule
    \rowcolor{black!8}\multicolumn{7}{l}{\textit{Noisy}} \\
    $-$\,heightmap-augmented critic & 95.0 & 89.5 & \underline{99.4} & 95.5 & 94.8 & 38.7 \\
    $-$\,WM-feature dropout         & \underline{97.5} & \underline{96.3} & \textbf{99.6} & \underline{96.7} & \underline{97.5} & \textbf{40.9} \\
    \cmidrule(l){1-7}
    \textbf{Ours}                   & \textbf{98.0} & \textbf{97.2} & 99.1 & \textbf{97.2} & \textbf{97.9} & \underline{40.6} \\
    \bottomrule
  \end{tabular}
\end{table}

\subsection{Full-Cover Stair-to-Plane Recovery Case Study}
\label{app:ood_dropout}

\begin{figure}[!t]
  \centering
  \includegraphics[width=\linewidth]{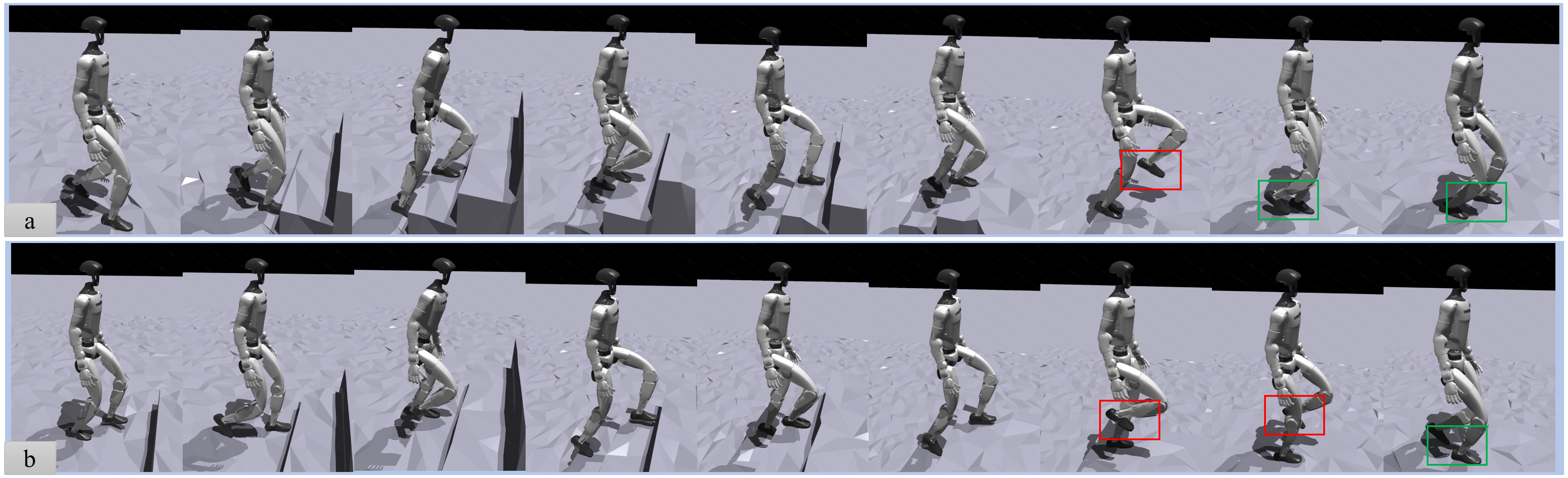}
  \caption{\textbf{Full-cover stair-to-plane recovery.}
  Stroboscopic rollouts on the same four-step
  ascending-stair-to-plane transition under full camera cover. Red boxes
  mark unnecessary stair-like high-clearance steps after the robot
  reaches flat ground; green boxes mark recovery to a steady plane
  stride. \textbf{(a)}~With WM-feature dropout, \method{} returns to
  plane walking after one extra clearance. \textbf{(b)}~Without
  WM-feature dropout, the policy continues stair-like high clearance for
  two steps before recovery.}
  \label{fig:ood_case}
\end{figure}

Removing WM-feature dropout has only a small effect in the Clean and
Noisy anchor cells (Appendix~\ref{app:ablations}), so those anchors do
not expose the regime targeted by this regularizer. We use a
qualitative full-cover case study to examine recovery when depth is
unavailable at a stair-to-plane transition. The robot climbs a
four-step ascending staircase with the camera fully covered and then
steps from the final stair onto flat ground.

Because depth is unavailable at the transition, the WM cannot
immediately correct the recurrent state from the new visual
observation. The policy may therefore receive a WM feature consistent
with stair climbing even after the ground has become flat.
Fig.~\ref{fig:ood_case} shows the behavioral consequence: with
WM-feature dropout, \method{} absorbs this stair-like guidance after one
extra high-clearance step, whereas removing dropout leaves the policy
in stair-like foot timing for two steps before it recovers. This
contrast supports the intended role of WM-feature dropout in the
WM overshoot regime of Section~\ref{sec:method-encoders}: under full
cover, the dropout-trained policy returns to plane walking sooner when
the WM feature remains biased toward stair climbing. This suggests that
dropout regularizes the policy against over-reliance on erroneous WM
features. The case targets full-cover recovery rather than normal
perceptive generalization; under clean or partially occluded
perception, depth remains available for the WM to adapt across the
transition.

\section{Real-World Experiments}
\label{app:real_details}

\begin{figure}[!t]
  \centering
  \includegraphics[width=\linewidth]{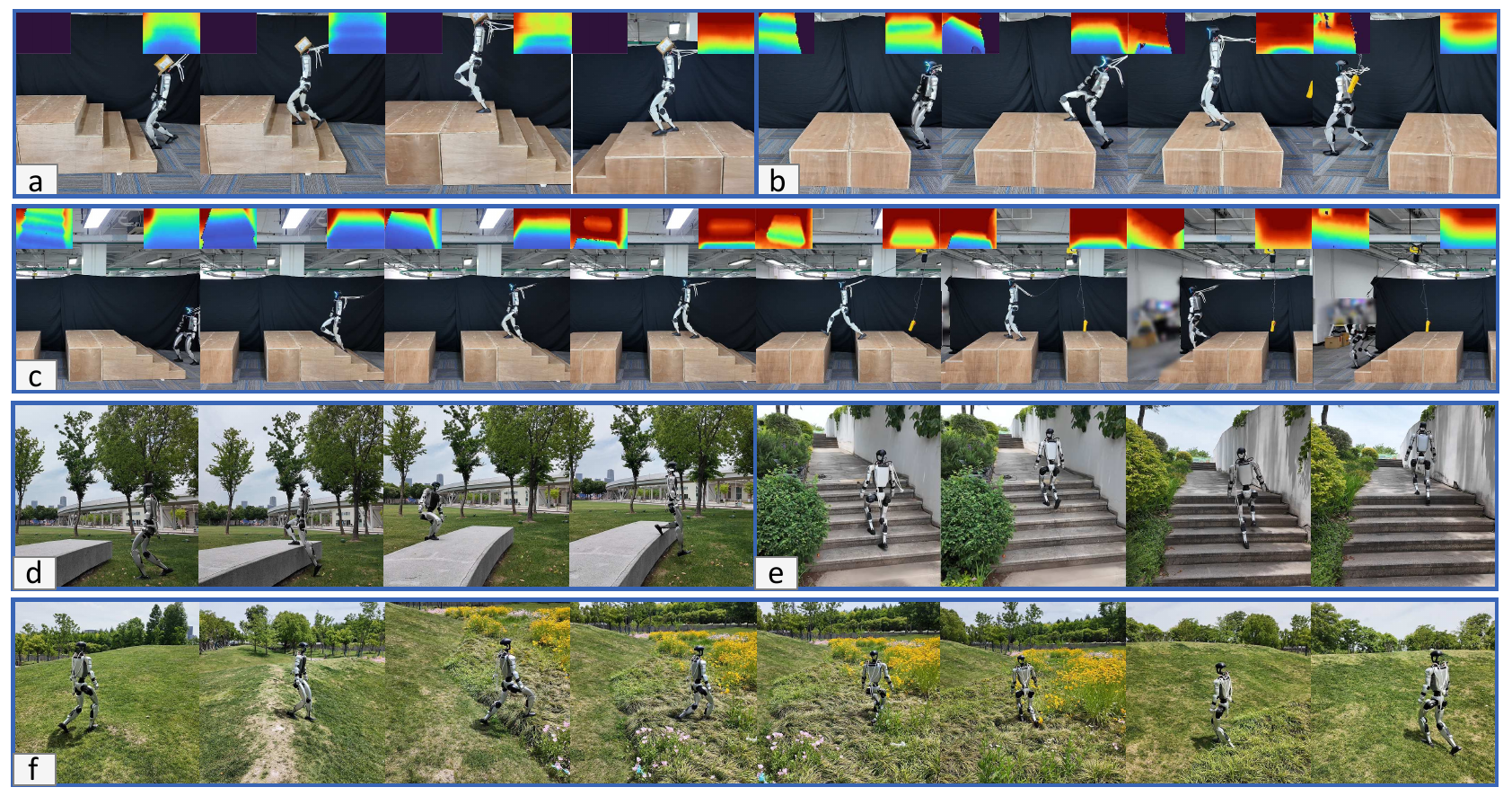}
  \caption{\textbf{Additional real-world deployment episodes.}
    Indoor controlled trials with raw-depth insets:
    (a)~stair traversal, (b)~platform traversal, and (c)~mixed-terrain
    traversal under perception perturbations. Outdoor deployment beyond
    the simulator's terrain and noise distribution:
    (d)~platform-edge traversal, (e)~stair traversal, and
    (f)~grass/vegetation traversal with depth artifacts.}
  \label{fig:real_extra}
\end{figure}

\subsection{Real-World Deployment Setup}
\label{app:real_setup}

We deploy \method{} on a Unitree G1 humanoid with a forward-facing
Intel RealSense D435i depth camera, running all perception and control
onboard on the robot's NVIDIA Jetson Orin module. The policy is
transferred directly from simulation with no real-world fine-tuning, and
controls only the lower body: it outputs 12 joint commands for the two
legs, covering hip pitch/roll/yaw, knee, ankle pitch, and ankle roll on
each side. The remaining waist and arm joints are held at fixed position
targets. The low-level interface uses Unitree's DDS command/state
channels: a 500\,Hz control loop publishes the latest PD joint targets,
while the learned policy updates the lower-body targets at 50\,Hz.

The depth path intentionally keeps real-camera preprocessing simple. The
D435i stream is read at 480$\times$270 resolution and converted to the
48$\times$64 depth tensor used by the WM through clipping to the
training depth range, resizing/cropping to the network input size, and
normalization. Unlike perceptive policies that rely on a tuned
deployment-time camera pipeline, we do not add real-specific depth
completion, hole filling, or temporal denoising at deployment: the same
lightweight geometric preprocessing is used for clean, occluded, and
flash-corrupted episodes, and residual real-sensor artifacts are handled
by the trained WM--policy system.

To preserve policy-side real-time execution, deployment separates the
low-rate WM update from the high-rate policy graph, which contains the
VAE encoder and actor.
The WM loop runs at 10\,Hz, consuming proprioception, the latest depth
frame, and recent actions to update its recurrent latent. After each WM
step, the deterministic latent is copied into a thread-safe shared
buffer. In parallel, the 50\,Hz policy loop reads the latest available
WM latent together with the current proprioceptive observation history.
Both models are exported as ONNX graphs and executed with ONNX Runtime;
preallocated input buffers and short, non-nested locks keep the
policy-side update latency at approximately 2\,ms, so occasional WM
updates do not stall high-rate action production.

\subsection{Additional Real-World Results}
\label{app:real_extra}

Fig.~\ref{fig:real_extra} collects additional real-world deployment
episodes beyond Fig.~\ref{fig:real}. The indoor panels show controlled
stair, platform, and mixed-terrain trials with synchronized raw-depth
insets, while the outdoor panels show transfer to platform edges,
outdoor stairs, and grass/vegetation scenes with depth artifacts. The
supplementary video shows the corresponding full episodes.

\end{document}